\documentclass{article}

\usepackage{times}
\usepackage{graphicx} 
\usepackage{subfigure} 

\usepackage{natbib}

\usepackage{algorithm}
\usepackage{algorithmic}

\usepackage{hyperref}

\usepackage[accepted]{icml2017}

\usepackage{anyfontsize}

\icmltitlerunning{RandomOut: Using a convolutional gradient norm to rescue
convolutional filters}

\begin{document} 

\twocolumn[
\icmltitle{RandomOut: Using a convolutional gradient norm\\to rescue
convolutional filters}



\icmlsetsymbol{equal}{*}

\begin{icmlauthorlist}
\icmlauthor{Joseph Paul Cohen}{udem}
\icmlauthor{Henry Z. Lo}{umb}
\icmlauthor{Wei Ding}{umb}
\end{icmlauthorlist}

\icmlaffiliation{udem}{Montreal Institute for Learning Algorithms (MILA), Universite de Montr\'{e}al}
\icmlaffiliation{umb}{University of Massachusetts Boston}
%
\icmlcorrespondingauthor{Joseph Paul Cohen}{cohenjos@iro.umontreal.ca}

\icmlkeywords{convolutional neural networks, regularization}

\vskip 0.3in
]



\printAffiliationsAndNotice{}  

\begin{abstract} 
Filters in convolutional neural networks are sensitive to their initialization. The random numbers used to initialize filters are a bias and determine if you will ``win" and converge to a satisfactory local minimum so we call this The Filter Lottery. We observe that the 28x28 Inception-V3 model without Batch Normalization fails to train 26\% of the time when varying the random seed alone. This is a problem that affects the trial and error process of designing a network. Because random seeds have a large impact it makes it hard to evaluate a network design without trying many different random starting weights. This work aims to reduce the bias imposed by the initial weights so a network converges more consistently. We propose to evaluate and replace specific convolutional filters that have little impact on the prediction. We use the gradient norm to evaluate the impact of a filter on error, and re-initialize filters when the gradient norm of its weights falls below a specific threshold.  This consistently improves accuracy on the 28x28 Inception-V3 with a median increase of +3.3\%. In effect our method RandomOut increases the number of filters explored without increasing the size of the network. We observe that the RandomOut method has more consistent generalization performance, having a standard deviation of 1.3\% instead of 2\% when varying random seeds, and does so faster and with fewer parameters.
\end{abstract}

\section{Introduction}

\begin{figure}
  \begin{center}
    \includegraphics[width=1.0\columnwidth]{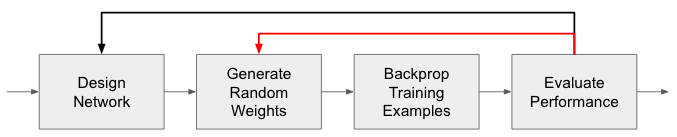}
     \caption{\textit{The Filter Lottery}. This is a problem that affects the trail and error process of designing a network. Because random seeds have a large impact it makes it hard to evaluate a network design without trying many different random starting weights. This work aims to reduce the bias imposed by the initial weights so a network converges more consistently. 
     }
      \label{fig:pipeline}
  \end{center}
\end{figure}

In convolutional neural networks \cite{lecun_convolutional_1995,lecun_deep_2015} different random seeds ({\em ceteris paribus}) greatly affect both the quality of the learned convolutional filters as measured by generalization error on the testing set.  We call this issue \textit{The Filter Lottery} because the random numbers used to initialize the network determine if you will ``win'' and converge to a satisfactory test error. The issue was mentioned in \cite{LeCun1998} and continues to be a challenge when training deep models which results in the typical workflow shown in Figure \ref{fig:pipeline}. In this work we explore it with a concrete example and propose a solution.

By simply changing the random initialization seed of a model we observe high variation in testing accuracy.  For example a 28x28 Inception-V3 model without Batch Normalization trained on CIFAR-10 fails 26\% of the time with an error was as low as random chance \cite{szegedy_rethinking_2015,krizhevsky_learning_2009}. The same phenomena was observed 5\% of the time when training a compact CraterCNN network on a dataset of Martian crater images \cite{cohen_crater_2016,bandeira_automatic_2010,cohen_randomout:_2016}. These results are to be expected because we are minimizing a non-convex loss function which we expect to have many local minima or saddle points that cause convergence behavior similar to that of local minima \cite{dauphin_identifying_2014}. 

\begin{figure*}
  \begin{center}
    \includegraphics[width=0.49\textwidth]{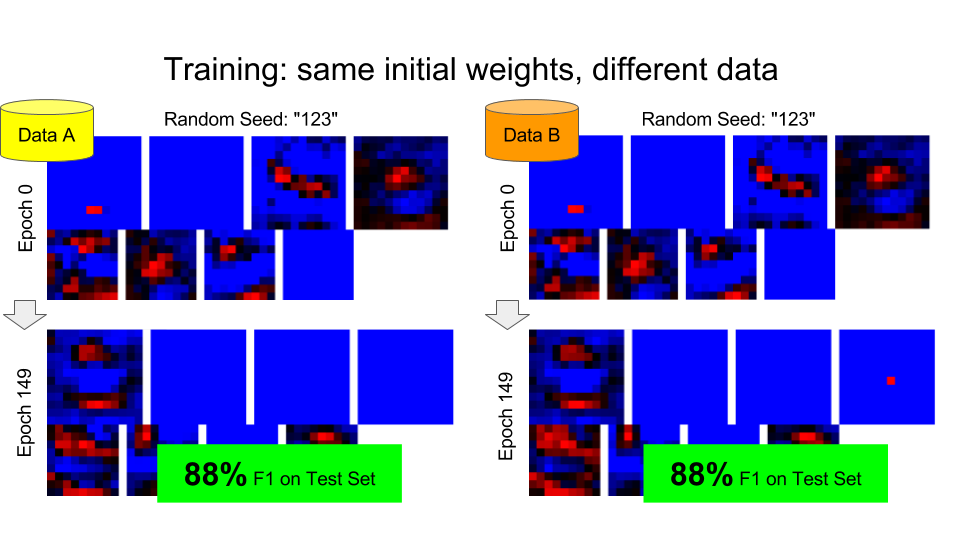}%
    \includegraphics[width=0.49\textwidth]{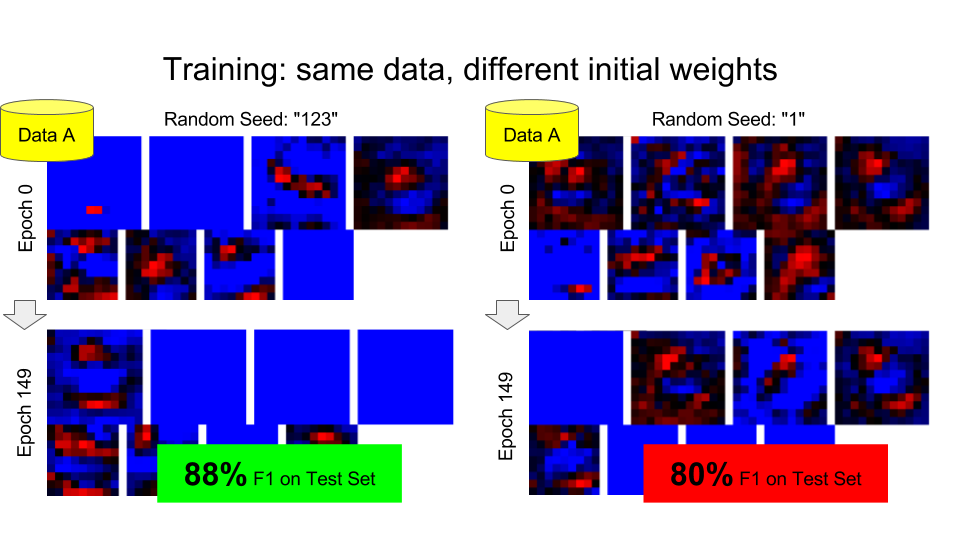}
     \caption{When training the CraterCNN on different examples but using the same random seed the network learned almost identical filters.
     When we vary random seeds and train on the same data we find that the filters learned are drastically different.
     }
      \label{fig:seed}
  \end{center}
\end{figure*}

We suspect this is due to the network not constructing the filters needed to extract the most discriminative features. Figure \ref{fig:seed} shows that varying the random seed can change how the filters will converge and also a large bias imposed by the random seed itself. When training the CraterCNN on different examples but the same random seed, the network learned almost identical filters. However, training on the same examples but with a different random seed the learned filters are drastically different, which has a strong impact on the testing accuracy.  This indicates that bad random seeds inhibit the learning of useful filters.  We call these filters ``abandoned'' by the network because they contribute little to minimizing the error.

Our experiments in \S \ref{sec:exp} indicate that wider networks (with more filters) have better performance. We believe that this is because more filters allows the network to successfully capture more discriminative features (similar to buying more lottery tickets). This would be unnecessary if all filters were utilized instead of being abandoned.

Carefully scaled initialization \cite{glorot_understanding_2010} and better optimization is not sufficient to solve this problem; the network may still start with bad filters. The random weights (not just the distribution they are drawn from) have an impact on the potential accuracy of the network. This problem is a result of the iterative methods used to train neural networks. Adding Batch Normalization \cite{Ioffe2015} layers resolve the problem in almost all cases but incur added runtime costs and parameters required after the training. These added layers lead to slower forward and back propagation because they are blocking operations that delay the next layers of the computation graph from processing. It is desirable to produce a model with minimum depth at test time and offload any added cost to training time.

We propose the method called \textsc{RandomOut} in \S \ref{sec:ro} that scores filters and replaces them at training time if they have been abandoned by the network. This can be thought of as a regularizer for convolutional filters to keep their gradients high. If a filter's contribution to the objective is insignificant, then we re-initialize it with random values and continue learning. This allows the weights to learn a completely different filter which will give the network another chance to reach an acceptable stationary point. \textsc{RandomOut} allows us to increase the number of filters explored without increasing the size of the network. This can potentially yield more compact networks which can train and predict with less computation.

\section{RandomOut}
\label{sec:ro}

	What information can the gradient contain? The derivative for each convolutional filter weight $\frac{\partial L}{\partial w_i}$ gives insight into the potential for that filter to influence the output. Here each $w_i$ is a weight in the network with a differentiable path to a loss function $L$. During backprop this value is calculated and used to determine its contribution to the output of the loss function. It is calculated by identifying the unique path of operations that are applied to $w_i$ and multiplying the local gradient at every operation. 
	
	Convolutional layers are no different in terms of the backprop algorithm. However for each convolutional filter we can calculate a holistic representation of it's influence to the loss function. We define the Convolutional Gradient Norm (CGN) in order to evaluate how much a filter (Lets call it $k$) will change as the network learns:
$$CGN(k) = \sum_{i} \left|\frac{\partial L}{\partial w^k_i}\right|$$%
We calculate this with respect to each minibatch. When the network error $L$ is low then it is expected that gradients will be low because we have converged. A low CGN and a low overall error implies a filter was learned correctly. However, a low CGN and a high overall error can imply the filter:

\vspace{-10pt}
\begin{enumerate}
  \item was abandoned by the network, decreasing its influence on the prediction in order to reduce error.
  \item overfit the training batch and learned some artifact of the training data. 
  \item extracts a useful feature but cannot be adjusted to correct the current erroneous predictions.
\end{enumerate}

\begin{figure*}
  \begin{center}
    \includegraphics[width=1\textwidth]{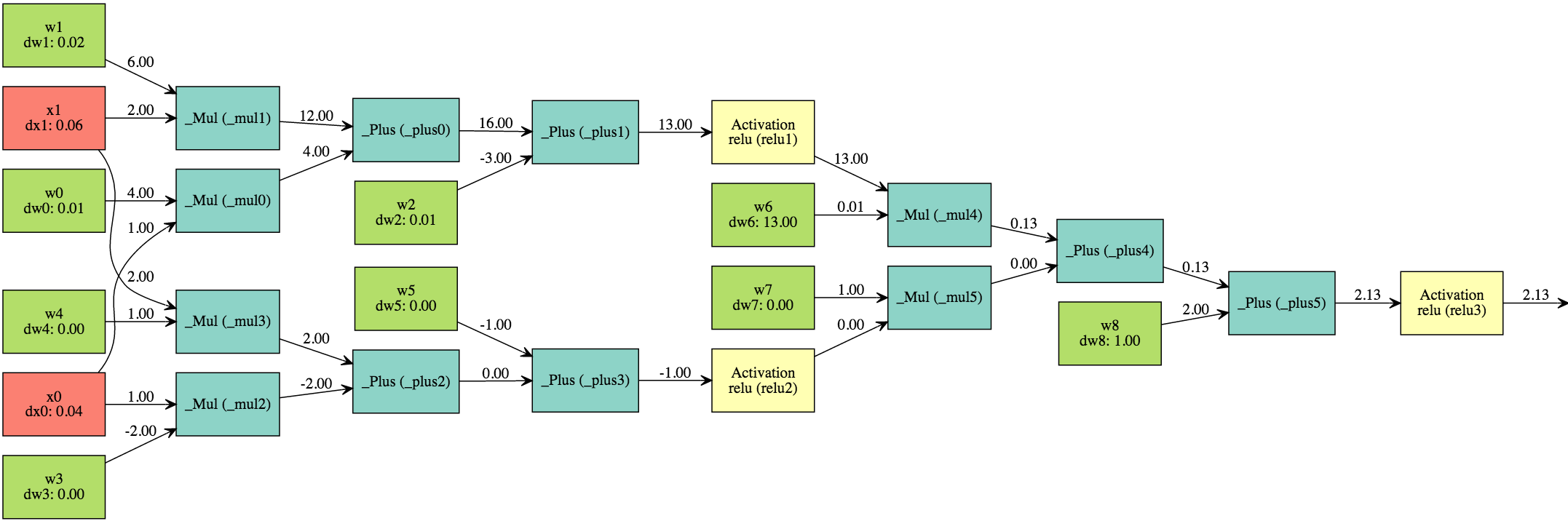}
    
    $f(x|w) = \max(0, w_6 \max(0,w_0x_0 + w_1x_1 + w_2) +  w_7 \max(0,w_3x_0 + w_4x_1 + w_5) + w_8)$
     \caption{The function $f(x|w)$ is decomposed into its computation graph. Edges represent the output of inputs and each intermediate computation. Gradients $\frac{\partial f}{\partial w}$ are shown inside the boxes of the inputs and weights.}
      \label{fig:compgraph-relu}
  \end{center}
\end{figure*}
	
What drives gradients to be low and parts of a computation graph to be abandoned? To explain the intuition we use the example in Figure \ref{fig:compgraph-relu} 
which shows a small neural network with ReLU activations. Here a gradient of 1 for $\frac{\partial f}{\partial f}$ is backpropagated down the network. Lets first focus on relu2. The output of plus3 is negative, so relu2 routes its incoming gradient to the 0 term of the $\max(0,-1)$ and a 0 gradient is routed to that portion of the network. This drives the gradients that directly depend on it ($\partial w_4$, $\partial w_3$, and $\partial w_5$) to be 0. These weights will never be updated unless some data in the future can cause the ReLU to activate. Alternatively, the weights $w_4, w_3$, and $w_5$ could be randomized to produce a better representation of $x_0$ and $x_1$ which could be more useful to minimize error.

We can also look at relu1 which is routing its gradient down the graph. The gradient applied to its downstream weights is low because the weight $w_6$ is very low, causing the gradient passed to relu1 to be low. This weight may be increased if $w_0, w_1$, and $w_2$ are transforming the inputs $x_0$ and $x_1$ into something that will minimize error. If not further training epochs would decrease their influence by reducing $w_6$ further.

The \textsc{RandomOut} approach is to reinitialize weights for abandoned parts of the network if their $CGN$ is below a threshold $\tau$ near 0. Because the gradients in these sections are low the impact is not very disruptive to the output. This is a concern because if the gradients to this region of the network were high, drastic changes to the weights would cause drastic changes to the output which would then cause large gradients to be sent back down the network which could cause havoc in the network. If the filters are randomized to a value that is used later in the network to reduce error its gradients will gradually increase and the section will slowly be introduced back into the network.

Formulating this into an algorithm, \textsc{RandomOut} has two hyperparameters, a threshold $\tau$ and a ``\% of epochs active'' $\mathcal{P}$. During training each filter $k$ is checked at regular intervals to see if $CGN(k) < \tau$ and, if so, filter $k$ is re-initialized. The motivation for the threshold is that the $CGN$ is hardly ever 0, because learning rates are fractional so update rules only approach 0, but will become very close when the network has stopped learning a filter. For our networks, $\tau = 10^{-8}$ yielded good results. It is also necessary to consider the proportion of epochs from the start in which filter randomization should occur; we refer to this as ``\% of epochs active'' or $\mathcal{P}$. Re-initializing filters too late in the training process may damage the network, and it will need time to retrain itself.

\section{Experimental Setup}
\label{sec:expsetup}

Two networks are used. The first is a small network used for filter visualizations, hyperparameter search, and testing capacity improvements called CraterCNN \cite{cohen_crater_2016} implemented in Deeplearning4j \cite{deeplearning4j_development_team_deeplearning4j:_2015} which is applied to Martian crater data. The second is a 28x28 Inception-V3 network which we use with and without Batch Normalization \cite{chen_mxnet:_2015} \cite{szegedy_rethinking_2015} from the MXNet repository. 


CraterCNN has two convolutional layers, followed by a fully connected layer, then softmax. The input is 15x15 and grayscale. Each crater candidate example is scaled to this size. The convolutional layers contain stride-1 4x4 filters.  The network uses ReLU activations as in \cite{krizhevsky_imagenet_2012}. The initial weights throughout the network are initialized using the Xavier initialization \cite{glorot_understanding_2010} scheme. It is trained using standard stochastic gradient descent with a fixed learning rate.

The Martian Crater dataset \cite{bandeira_automatic_2010} is used because it is challenging enough, while fast enough to perform the training of $>$14,000 networks ($>$2 million epochs) for hyperparameter tuning using grid search. We use three subsets of the data (the East, Center, and West regions) for our experiments which are split each region 50/50 into a training and test set. The East region contains 458/765 positive/negative crater examples and the West contains 1121/1385.

The 28x28 Inception-V3 network is used as implemented in MXNet example repository \cite{chen_mxnet:_2015} \cite{szegedy_rethinking_2015} and applied to CIFAR-10. It contains 6848 convolutional filters of sizes 3x3 and 1x1, trained using Adam \cite{kingma_adam:_2014}. Typically the network has Batch Normalization nodes after the output of every convolutional layer but we remove these layers to demonstrate \textsc{RandomOut} because the methods are not compatible. The 28x28 Inception-V3 network without Batch Normalization is referred to as Base and is referred to as BatchNorm when those layers are included.

Full code examples are available online\footnote{\url{https://github.com/ieee8023/NeuralNetwork-Examples/tree/master/mxnet/randomout}}

\section{Experiments}
\label{sec:exp}

\begin{figure}
  \begin{center}
    \includegraphics[width=1.0\columnwidth]{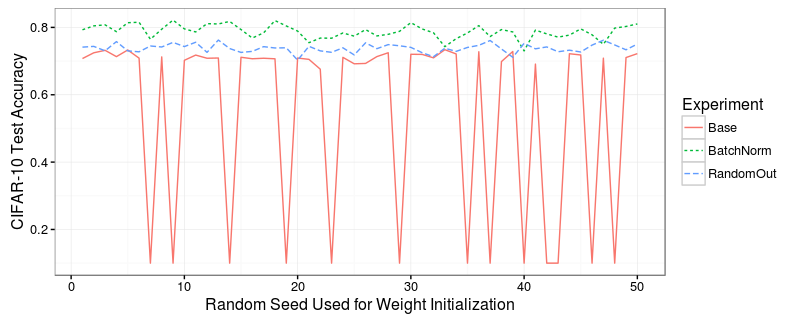}
     \caption{The testing accuracy of the network is plotted while varying nothing but the random seeds used to initialize the network.}
      \label{fig:varyseed}
  \end{center}
\end{figure}

Overall results are presented in Figure \ref{fig:varyseed}. The resulting test error of the 28x28 Inception-V3 network in three experimental conditions are shown. The base network fails to converge to a satisfactory local minimum $26\%$ of the time. We draw the reader's attention to the effectiveness of our method in recovering from all bad seeds and increasing accuracy overall. \textsc{RandomOut} allows the network to recover from being initialized with bad weights. However using BatchNorm appears to perform even better except for 3 random seeds where it is about equal. We also observe more consistency in generalization performance when using \textsc{RandomOut}. BatchNorm test accuracy has a standard deviation of 2\% while \textsc{RandomOut} is almost half at 1.3\%. 

\begin{figure}
    \centering
    \includegraphics[width=1\columnwidth]{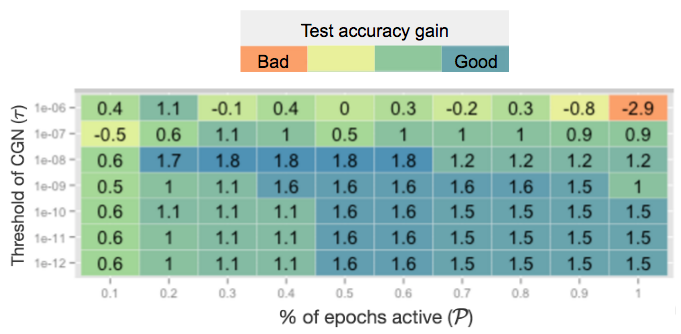}
    \caption{Resulting testing accuracy gain of when varying the two hyperparameters of \textsc{RandomOut}. The threshold $\tau$ and ``\%
     of epochs active'' $\mathcal{P}$ are varied. Each cell value in the heatmap is the mean gain of of 50 different random seeds when using \textsc{RandomOut}.}
    \label{fig:var}
\end{figure}

In order achieve these results we needed to determine how to set the hyperparameters $\tau$ and $\mathcal{P}$. In order to perform a hyperparameter search we used the smaller CraterCNN network and the small Crater dataset. We evaluate \textsc{RandomOut} on 50 random seeds used for initialization over 150 training epochs. Test accuracy when varying $\tau$ and $\mathcal{P}$ is shown in Figure \ref{fig:var}. We observe that generally for a fixed $\mathcal{P}$ a lower threshold $\tau$ value results in a higher average gain in network accuracy. This is because these filters have been correctly identified as being abandoned and brought back to life to improve the network. We also find that increasing $\mathcal{P}$, provided a low threshold $\tau$, yields a higher gain. We understand this to mean that the lower the threshold is the lower the risk of randomizing a filter that has learned an important feature. It is significant to note that these improvements are not brittle and just for specific hyperparameters but can be observed over large areas of the parameter spaces as shown in the green and blue cells in Figure \ref{fig:var}.

\begin{figure*}
    \centering
    \includegraphics[width=0.8\textwidth]{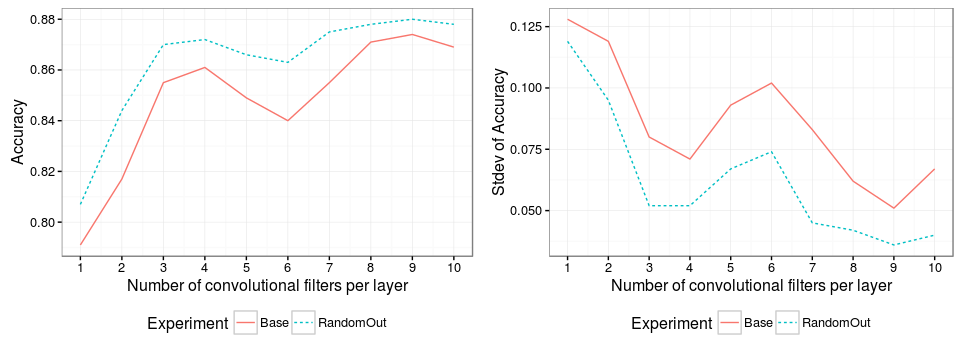}
    \caption{Here the number of filters used in the network is varied between 1 to 10 with and without \textsc{RandomOut} enabled. This plot shows the mean accuracy score from 50 different random seeds of CNNs with \textsc{RandomOut} lead those without it by around 1 to 2 filters consistently. This means the \textsc{RandomOut} method enables CNNs to increase their accuracy to that of a network containing more filters but without the computational cost of actually adding more filters.}
    \label{fig:varfilters}
\end{figure*}

Now we use the hyperparameters $\mathcal{P} = 1.0$ and $\tau = 10^{-12}$ and vary the number of filters used in each layer of the CraterCNN network configuration in Figure \ref{fig:varfilters}. The goal of this figure is to determine if we achieve an increase in the potential of a network to that of one with more filters without incurring the cost of adding more filters. This network is very small with only two convolutional layers so a 1 in this plot means only 1 filter in each layer. The mean accuracy is used from 50 networks each using a different random seed and trained for 100 full size batch epochs. 

We can observe the performance CNNs with \textsc{RandomOut} lead those without it by around 1 to 2 filters consistently. These results indicate that \textsc{RandomOut} achieves an increase in performance equivalent to about 1 or 2 added filters for this network. For example on the West region using 1 filter with \textsc{RandomOut} achieves the same performance of using 2 filters without it. We can also observe a smoothing effect on the accuracy at 6 filters in every region. It is unknown what caused this dip in performance but it is clear that \textsc{RandomOut} mitigated this negative effect. These finding support our statement that these networks have abandoned filters because randomizing them achieves similar performance to making them wider.

\begin{figure}
\centering
\includegraphics[width=0.9\columnwidth]{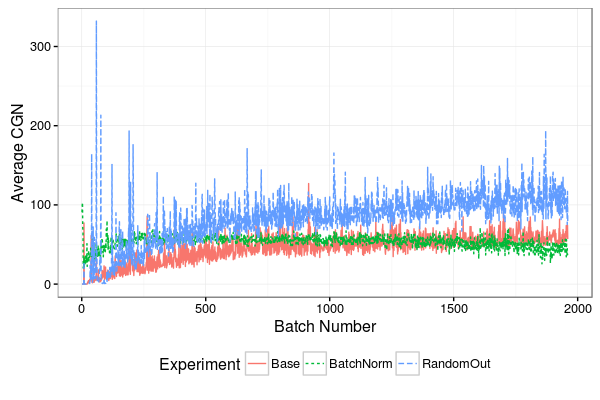}
    \caption{During each batch the CGN is calculated for every filter. Here the average CGN is shown when training the 28x28 inception-v3 network in three test conditions. The average is taken for all 6848 filters in the network.}
    \label{fig:batch-cgn}
\end{figure}

\begin{figure}
\includegraphics[width=0.9\columnwidth]{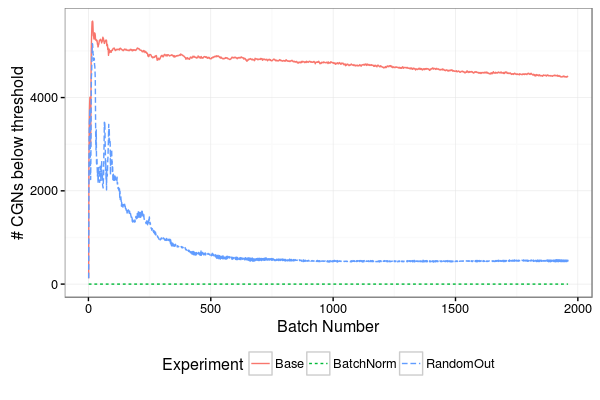}
    \caption{During training every epoch each filter's CGN is compared to the threshold. The number of CGNs that are below the threshold (and would be reinitialized by the \textsc{RandomOut} algorithm) are shown.}
    \label{fig:batch-belowthresh}
\end{figure}

Next we go deeper into how \textsc{RandomOut} is performing during training. In Figure \ref{fig:batch-cgn} the average CGN is shown when training the 28x28 inception-v3 network in three test conditions. \textsc{RandomOut} continuously increases the gradients while BatchNorm constrains them. The large spikes in the beginning can be explained by Figure \ref{fig:batch-belowthresh} which shows the number of filters that fall below $\tau$. This represents how many are reinitialized by \textsc{RandomOut} and how many would be reinitialized for the base and BatchNorm lines. There are a large number of resets at the start of training and then this decreases quickly as training continues.

\begin{figure*}
\centering
    \includegraphics[width=0.8\textwidth]{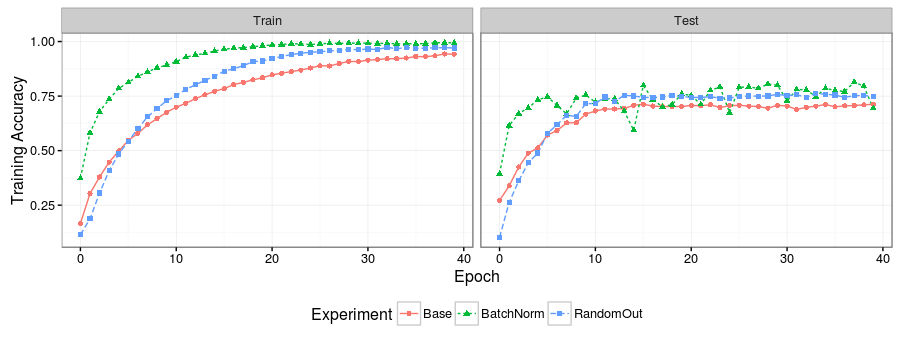}
\caption{The training and testing accuracy of the 28x28 inception are shown here. The same initial seed is used for all networks so they all start with the same weights. Here it can be seen that the training error of \textsc{RandomOut} is more consistant than BatchNorm.}
\label{fig:epoch-acc}
\end{figure*}

We can then look at the training and testing accuracy of the 28x28 Inception model under the three test conditions in Figure \ref{fig:epoch-acc}. Using BatchNorm causes testing accuracy to be unstable while at times achieving the highest accuracy. BatchNorm accelerates training a reaches a higher training and testing accuracy faster. \textsc{RandomOut} is slower to surpass the Base condition but achieves a consistent gain in testing accuracy. We draw the reader's attention to the stable training and testing accuracy while \textsc{RandomOut} is continually reinitializing filter weights.

\section{Conclusion}

We introduced \textsc{RandomOut}, a method of detecting and repairing abandoned conv. filters with the goal of reducing the bias by initial weights. We analyse the causes of the issue and hyperparameters on examples and then demonstrate an increase in performance on the well known CIFAR-10 dataset and the 28x28 Inception model. We conclude that the hyperparameters are easy to config. with $\mathcal{P}$ = 1 and $\tau$ very close to 0.

Although the addition of BatchNorm layers yields higher accuracy, it incurs added runtime costs and parameters, unlike in \textsc{RandomOut}. We study the effect of RandomOut when varying the number of filters in a network and conclude that RandomOut increases accuracy of a network consistently. We study the CGN value with RandomOut and BatchNorm at every batch to find that these methods produce drastically different patterns of gradients while training which demonstrates the difficulty in merging these two methods. We finally demonstrate the stability of testing accuracy on models produced with \textsc{RandomOut} compared to BatchNorm. We expected that resetting weights might damage the network but discovered that because we are resetting weights with small gradients the impact on the network is low.

{\fontsize{9}{9}\selectfont
\bibliography{neuralnetworks,neuralnetworks2,joe,craters} 
\bibliographystyle{icml2017}
}

\end{document}